\documentclass[11pt]{article}
\usepackage{coling2020}
\usepackage{times}
\usepackage{url}
\usepackage{latexsym}
\colingfinalcopy
\usepackage{xcolor}
\usepackage{soul}
\renewcommand{\hl}[1]{#1}
\usepackage{graphbox}
\usepackage{versions}\excludeversion{ignore}
\usepackage{amsmath}
\usepackage{graphicx}
\usepackage{cjhebrew}
\usepackage{arabtex}
\usepackage{subfig}
\newcommand{\myparagraph}[1]{\textbf{#1}}  
\renewcommand{\subsubsection}[1]{\myparagraph{#1.}}
\title{Transliteration of Judeo-Arabic Texts into Arabic Script \\Using Recurrent Neural Networks}
\author{Ori Terner \\
  School of Computer Science\\ 
  Tel Aviv University\\
  Ramat Aviv, Israel\\
  {\tt oriterner@gmail.com} \\\And
  Kfir Bar \\
  School of Computer Science \\
  College of Management \\
  Academic Studies \\
  Rishon LeZion, Israel \\
  {\tt kfirb@colman.ac.il} \\\And
  Nachum Dershowitz \\
  School of Computer Science\\ 
  Tel Aviv University\\
  Ramat Aviv, Israel\\
  {\tt nachum@tau.ac.il} \\}

\date{}
\begin{document}
\maketitle
\begin{abstract}
We trained a  model to automatically transliterate Judeo-Arabic texts into Arabic script, enabling Arabic readers to access those writings. 
We employ a recurrent neural network (RNN), combined with the connectionist temporal classification (CTC) loss to deal with unequal input/output lengths.
This  obligates adjustments in the training data to avoid input sequences that are shorter than their corresponding outputs. 
We also utilize a pretraining stage with a different loss function to improve  network converge. 
Since only a single source of parallel text was available for training, we take advantage of  the possibility of generating data synthetically.
We train a model that has the capability to memorize words in the output language, and that also utilizes  context for distinguishing ambiguities in the transliteration. 
We obtain an improvement over the baseline 9.5\% character error, achieving 2\% error with our best configuration.
To measure the  contribution of context to learning, we also   tested  word-shuffled data, for which the error rises to 2.5\%.
\end{list} 
\end{abstract}

\section{Introduction}

Many great Jewish literary works of the Middle Ages were written in Judeo-Arabic, a Jewish dialect of the Arabic language family that adopts the Hebrew script as its writing system. 
Prominent authors include Maimonides (12th c.), Judah Halevi (11th--12th c.), and Saadia Gaon (10th c.).
\begin{ignore}
This has several reasons, as Judah ben Saul ibn Tibbon mentions in his preface to his Hebrew translation of the book \textit{Al Hidayah ila Faraid al-\d{K}ulub [Direction to the Duties of the Heart]} by {Bahya ibn Paquda}, which was also written in this manner.
\begin{quote}
First, most of the \textit{Geonim} in Babylon and in the region of Israel and the Persian Empire were living in Arab speaking countries, and were speaking the Arabic language. They did so [writing in Arabic] because all the people understood this language. Secondly, since it is broad by any means, and it is sufficient for every speaker and writer, and the rhetoric is straight and clear, reaching the objective in every subject more than is possible with the Hebrew language. This is because in the Hebrew language we don’t have but what is found in the books of the Bible, and it is not sufficient for every need of a speaker. Also their aim was to benefit, in their writings, uneducated people who were not proficient in Hebrew [though they did know the alphabet].\footnote{Translated from the source. See {\small\url{http://www.daat.ac.il/daat/vl/hovatlevavot/hovatlevavot06.pdf}}.}
\end{quote}

This phenomenon of transcribing the spoken language with Hebrew scripts among Jewish communities is not exclusive for Jews living in Arab speaking regions. 
Other notable examples are:
\begin{itemize}
    \item \textit{Loazei Rashi} -- Rashi (Rabbi Shlomo Yitzchaki) in his commentary on the Talmud and the Bible occasionally explains a term by giving its translation in Old French, his vernacular language, transcribed in Hebrew script. There are thousands of such examples, giving latter-day scholars a window into the vocabulary and pronunciation of Old French \cite{10.2307/435416}.
    \item
    A second example is Yiddish, the historical language of the Ashkenazi Jews. {Yiddish} is a Judeo-German language, and it is written in the Hebrew alphabet. Unlike the first example, {Yiddish} is transcribed as full sentences, not just single words, allowing context and syntax to be examined. {Yiddish} is still in use in the present day in certain communities as a vernacular.
    \item
    Also {Ladino}, the Judeo-Spanish Romance language, was formerly written with Hebrew script.
\end{itemize}
It is presumed that this use of Hebrew script instead of the Arabic one is not solely a matter of convenience, but also in certain cases a matter of discretion, that is, placing a barrier to discourage non-Jewish readers. 
\end{ignore}
In this work, we develop an automatic transliteration system that converts Hebrew-letter Judeo-Arabic into readable Arabic text. 

\begin{ignore}
The task is that of generating, from the original Judeo-Arabic text, a sequence of Arabic letters that corresponds to the word sequence \emph{intended} by the author of the text. 
\end{ignore}


\begin{ignore}
There is a distinction between two terms, \textit{transliteration} and \textit{transcription}, both constitute the transformation of text written in the script of one language into the script of another. While transliteration is a process of changing each grapheme in the source language to a grapheme in the target language, usually in a one-to-one manner, transcription is a process that seeks to preserve the phonemes, that is the way the original words sound.

The border between the two terms is not always definite. In the matter under discussion, that is Judeo-Arabic, the classification tends more to the transliteration side. 
\end{ignore}

Generally speaking, given a text, transliteration is a process of converting the original graphemes to a sequence of graphemes in a target script. 
Specifically, transliterating a Judeo-Arabic text into the Arabic script almost invariably results in a text that has a similar number of letters. Yet, the correspondence between the letters in the transliteration is not one to one. 
 Judeo-Arabic Hebrew script includes matres lectionis \smash{(e.g.\@ \RL{A, w, y } )} to mark some of the vowels but it typically does not include nunation -- \textit{tanween} in Arabic \smash{(e.g.\@ \fullvocalize
 \RL{baN} \novocalize)}. Additionally, the \textit{hamza} letter \smash{(\RL{"'})}, a relative latecomer to the Arabic writing system \cite{Shaddel2018-SHATOT-7}, is missing in the Judeo-Arabic script when it is placed {``on the line''} and not as a decoration for one of the matres lectionis. 
 
\begin{figure}[t!]
  \centering
  \subfloat[Left: manuscript, right: printed edition, but digital-text reads: \cjRL{nyt'} (missing \textit{shadda} diacritic)] {\includegraphics[width=0.44\textwidth]{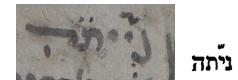}}
\qquad 
  \subfloat[Left: manuscript, right: printed edition, but digital-text reads: \cjRL{n'q.s'} (missing \textit{tanwin} diacritic)]{\includegraphics[width=0.45\textwidth]{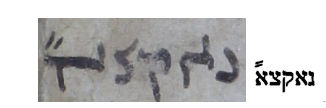}}
  \caption[Information missing from the digital text]{Information missing from the digital text and present in the manuscript and critical edition.}
  \label{fig:mis_inf}
\end{figure}

 Some other challenges are:
(1) Authors of Judeo-Arabic texts sometimes use different mappings between Hebrew and Arabic letters. 
Some authors use the Hebrew letter \cjRL{g} to transliterate the Arabic letter \smash{\RL{j}}, and others will use it to transliterate the Arabic letter \smash{\RL{.g}}.
(2) Diacritic marks (small dots placed either above or below letters) are often omitted in the Hebrew script, another source of ambiguity. 
For example, the Arabic letters \smash{\RL{d}} and \smash{\RL{_d}} are sometimes represented by the same letter in Hebrew. 
When the diacritic marks are maintained in the original Hebrew script, usually they are used in an inconsistent way. 
Even if they are used in the original manuscript, in many cases those marks appear differently or are completely missing in  digital editions (such as those  we used in this work). 
Figure \ref{fig:mis_inf} shows a few examples of this problem.
The \textit{apostrophe} is the only diacritic mark that is  used in the digital texts  we used. 
One important diacritic mark that is often missing from digital versions is \textit{shadda} (gemination), which may be used to easily resolve some lexical ambiguities. 
For example, the word \smash{\RL{drs}} (``he studied'') has a different meaning when it appears with a \textit{shadda} on the second consonant \smash{\RL{drrs}} (``he taught'').
(3) There are several Arabic letters that do not have a one-to-one mapping with a Hebrew letter. 
For example, the Hebrew letter \smash{\cjRL{y}} typically refers  either to \smash{\RL{y} or \RL{Y}} when  appearing at the end of a word.
(4) Judeo-Arabic is heavily affected by Hebrew and Aramaic; therefore, Judeo-Arabic texts are typically enriched with Hebrew and Aramaic citations, as well as with borrowed words and expressions from the two languages. 
Those citations and borrowings should not be blithely transliterated into Arabic, but rather need to be either left in the original script, semantically translated into Arabic, or otherwise annotated. 
Sometimes those borrowed words get inflected as if they were original Arabic words. For example, the word \smash{\cjRL{'l/skynh}}, composed of the Hebrew  \smash{\cjRL{/skynh}} (\textit{shkhina}, “divine spirit”) and  Arabic definite article \smash{\cjRL{'l}} (in Arabic \smash{\RL{Al}}); see \cite{inproceedings}.

The Friedberg Jewish Manuscript Society (\url{https://fjms.genizah.org}) has recently released a collection of hundreds of Judeo-Arabic works from different periods of time, formatted digitally as plain text and decorated with HTML tags. Hebrew and Aramaic citations and borrowings are annotated in those texts by domain experts. 
In the present work, we use texts from this collection. 
We noticed that some common borrowings were missed by the annotators; therefore, in this work we focus on the task of the transliteration of Judeo-Arabic words that originated in Arabic words only. 
Detecting the boundaries between Arabic-origin words and borrowings from other languages, a task that is  known as ``code switching'', is left for future improvements. Since in our approach we consider the context of a word that needs to be transliterated, mostly to resolve lexical ambiguity, we masked those citation and borrowings, allowing the model to handle a continuous sequence of characters.

\begin{ignore}
In this work, we focus did not cover this aspect in this work. 
Instead we replaced all {non-Arabic words} in the data with a reserved symbol.
\end{ignore}

This Judeo-Arabic-to-Arabic transliteration problem has been addressed in one previous work \cite{inproceedings}.
We elaborate on this and some other relevant previous works in the next section. 
In this work, we propose an end-to-end model to handle the transliteration task, by training a flavor of a recurrent neural network (RNN) 
on relatively short parallel texts, augmented with some synthetically generated data. 
Our model was designed with the capability of memorizing words in the output language.
Section 3 describes the texts at our disposal.
It is followed by a description of the baseline algorithm,
and then our RNN solution.



\section{Related Work}

{The only previous effort dealing with the transliteration of} Judeo-Arabic texts \cite{inproceedings} employed a method  inspired by  statistical machine translation,  which was state of the art before {deep neural nets} took over. 
It consisted of a log linear model where the main component is a phrase table that counts the number of occurrences in the training data. 
As in \cite{DBLP:journals/corr/RoscaB16}, they also regarded  transliteration as translation at the character level, that is, imagining single letters to be words. 
\hl{They improved their results by reranking their model's top predictions using a word-level language model. 
This is expected to be beneficial since using a character-level mechanism would incur the danger of generating  non-words and nonsensical sequences of letters.}
A word-level language model that is rich enough to avoid high {unknown} rates would screen those results out.

Proper noun transliteration is a related task, which has been addressed previously in many works. 
The input term is usually provided without context, and the task is to rewrite the name in the target script, maintaining the way the name is pronounced.
A direct phrase-based statistical approach to transliterating proper nouns from English to Arabic script is described in \cite{Nasreen}.
It was preceded by largely handcrafted methods for romanization from Arabic to English such as \cite{Arbabi,Stalls}.
A deep-belief approach for proper names
was taken in \cite{10.5555/1626431.1626476}.
An attention-based method  for transliteration from Arabic named entities to English and vice-versa is taken in \cite{HADJAMEUR2017287}.

In \newcite{DBLP:journals/corr/RoscaB16}, a transliteration system of  names that employs a recurrent neural network (RNN), in two different ways, was described. 
In one way, they developed a sequence-to-sequence model that elegantly handles the different lengths of the input and output, with CTC alignment (see {Section \ref{CTC}}). 
A minor difference is their use of ``{epsilon insertion}'' to deal with input sequences that are shorter than the matching output, which CTC cannot handle. 
We instead use a similar solution of letter doubling discussed later.
The second approach is a model inspired by recent work in the field of machine translation applying an encoder-decoder architecture with an attention mechanism. 
They report on improved results using RNN compared to previous methods, as is usually the case when employing deep learning techniques with problem previously solved by other means.

Another  closely related problem is the transcription of \textit{Arabizi}, which is a romanized transcription of Arabic that emerged naturally as a means for writing Arabic in chats,
especially in the days of limited keyboards. 
Transcription  into Arabic graphemes was studied by \cite{bies2014transliteration}. 
Except for the different dialects and {code switches} from colloquial Arabic of different dialects to modern standard Arabic (MSA), they also had to handle changes in the language induced by the platform it appears in, including \textit{emoticons} and deliberate manipulation of words, such as repetition of a single letter for emphasis common with internet social media.
The same problem {was addressed in} \cite{al2014automatic}, which  maps  Roman letters to Arabic scripts to produce a set of possibilities and chooses from them  using a language model. 

In another related task, the goal was to translate written text to the International Phonetic Alphabet (IPA),
capturing the pronunciation of the written text. 
{The work in} \cite{rao2015grapheme} tried different RNN models, handling unequal sized input-output pairs by epsilon post-padding. They experimented with time delays, that is, postponing the output by a few timestamps (by pre-padding the output while post-padding the input accordingly). 
This allows the network to catch more of the input before deciding on the output. 
They also compared this with a bidirectional long-short-term-memory (LSTM) network that is able to see the entire context backward and forward.
They combined the bidirectional LSTM with a CTC layer, handling the longer-output-than-input issue with epsilon post-padding (we used {doubling}). 
They reported, as expected, that greater contextual information contributes to  performance. 
The bidirectional LSTM performs better than the unidirectional one even when the full context (whole word) is fed to the network before it starts its prediction. 
The best performance was obtained using the bidirectional LSTM, combined with an n-gram (non-RNN) model.

This transcription from a written language to IPA can potentially be used as a mediator to {transcribe} between any two languages were we to have an encoder from IPA to the source language and a decoder from IPA for the target script. 
Another potential use for this ability of transliteration to IPA would be for improved spell checking and correction (for those who only have basic knowledge of the sounds of the letters). 
A model could predict the utterances from the graphemes and recognize the word by similarity in the sound domain, to arrive at the correct spelling. \hl{For example:} \textit{neyber  $\rightarrow$ ne{\small I}ber  $\rightarrow$ neighbor}.



\section{Data}

\subsubsection{\textit{Kuzari}}  
To train and evaluate our RNN model, we needed a considerable amount of parallel text in Judeo-Arabic along with its Arabic transliteration. 
Thankfully, the \textit{Kuzari} (\textit{Kitab al Khazari}), a medieval philosophical treatise composed by Judah Halevi in Andalusia around 1140, was  recast 
in Arabic by {Nabih Bashir} \cite{Bashir}. 
We use this Arabic version and the original Judeo-Arabic as a parallel resource for training. 
The original Judeo-Arabic text of the Kuzari is taken from the {Friedberg repository}, and is derived from the critical edition edited by David H.\@ Baneth. 
It was based on several manuscripts and comprises 47,348 \hl{word tokens (15,532 unique types)}, about 11\% of which are Hebrew and Aramaic insertions. 
It is important to mention that Bashir's book provides translations for those insertions, and that there is no distinction in his edition between the \textit{transliterated} and \textit{translated} words. As mentioned before, we postpone handling the Hebrew and Aramaic citation and borrowings in our work, and we use the annotations provided by the Friedberg's experts in the Judeo-Arabic text, to mask out all words that are not originated in Arabic.

After cleaning the data from translation comments,   section numbering and other elements irrelevant for  transliteration, we needed to align the Arabic translation with the Judeo-Arabic source, on the word level. 
Aligning the texts is necessary because Bashir’s translation does not perfectly match the Judeo-Arabic text, there being some word insertions, deletions and edits.
To do that, we begin by breaking the two texts into words, and then iterating the two sequences of words. 
At each iteration we choose between skipping a word from the source (Judeo-Arabic) text, skipping a word from the target (Arabic) text, or adding both as an aligned source/target pair. 
The decision is  the choice that minimizes the total edit distance of the alignment. 
To calculate edit distance between strings of different scripts, we applied the simple map to the Arabic source to get a basic transliteration into Hebrew letters. 
{(The details of the algorithm are given in} \cite[Appendix]{Thesis}.)
\begin{ignore}
. \\ \\
For:\\
$T_i^{[0,...,l_i-1]}=\left(t_i^{0},t_i^{1},...,t_i^{l_i-1}\right)$ -- word sequence with length $l_i$\\
$\epsilon$ -- the empty string \\ 
$\phi$ -- an empty sequence \\ \\
Define:\\
$\textit{AlignScore}\left( T_1^{[0,...,l_1-1]},T_2^{[0,...,l_2-1]}\right)=$ 
\begin{align*}\label{eq}
=\begin{cases}
0, & \text{if } l_1=0 \wedge l_2=0  \\
\textit{AlignScore}\left(\phi,T_2^{[0,...,l_2-2]}\right) + \textit{editDist}\left(\epsilon,t_2^{l_2-1}\right), & \text{if } l_1=0\\
\textit{AlignScore}\left(T_1^{[0,...,l_1-2]},\phi\right) + \textit{editDist}\left(t_1^{l_1-1},\epsilon\right), & \text{if }l_2=0\\
\min \left\{\begin{array}{lr}
        \textit{AlignScore}\left(T_1^{[0,...,l_1-2]},T_2^{[0,...,l_1-2]}\right) + \textit{editDist}\left(t_1^{l_1-1},t_2^{l_2-1}\right),\\
        \textit{AlignScore}\left(T_1^{[0,...,l_1-2]},T_2^{[0,...,l_2-1]}\right) + \textit{editDist}\left(t_1^{l_1-1},\epsilon\right),\\
        \textit{AlignScore}\left(T_1^{[0,...,l_1-1]},T_2^{[0,...,l_1-2]}\right) + \textit{editDist}\left(\epsilon,t_2^{l_2-1}\right)
        \end{array}\right\}
 , & \text{otherwise}
\end{cases}
\end{align*}

Remarks:
(1)
Memoizing at each stage what choice was made by the \textit{minimum} operator, enables to track back the desired alignment.
To implement this algorithm in a non-space-consuming manner, only the alignment scores of two levels of recursion where kept. 
This is sufficient since the \textit{minimum} operator operates only one level deep. 
The actual implementation was  iterative.
(2) The $\textit{editDist}$ maps both strings to Judeo-Arabic according to a simple mapping (Table \ref{tab:simp_map} in Appendix), before calculating  distance.
(3)The $\textit{editDist}$ on the empty string is equivalent to the length of the other string.
\end{ignore}
After this alignment process we only keep pairs of words in the alignment that are close enough edit-distance wise. We set the threshold for edit-distance similarity heuristically, after normalizing the edit distance result by the maximum length of the two words that were compared, to the value of 0.5.
This left us with a total of 47,083 word pairs, {of which 20\%  were} chosen randomly as test data, and the rest left for training.
We are aware of the {generalization} problem of training and testing on data from the same source. 

Words that do not align well according to the predefined threshold, such as Hebrew insertions, are removed and replaced by a fixed symbol \hl{(H)} both in the source and target texts. We do the same for Hebrew insertions that are annotated as such by Friedberg's annotators.
We do this to preserve the sentence structure since we suspect that this  information is {of value to} the model for making better predictions.

We see punctuation as an important feature of the language and we want to keep this information for the model to train with. 
Where there are {disagreements} (e.g.\@ comma vs.\@ {period}) in the source and target languages, we favor the source language. 

Additionaly, we removed from the Arabic text the \textit{harakat} marks (short vowel marks): \textit{fathah}, \textit{kasrah}, \textit{dammah} and \textit{sukun}. They appear rarely in written Arabic (except for text intended for children or in religious texts) and are usually considered  ``noise'' for language-related algorithms due to their scarcity \cite{habash2010Arabic}. 
In the Judeo-Arabic text, the equivalent to these diacritics is the Hebrew \textit{niqqud} signs. 
They appear only for  Hebrew insertions; hence they are removed incidentally by the replacement of Hebrew insertions. 
Sometimes we even make use of those marks to identify Hebrew insertions missed by Friedberg's experts.
Note that the \textit{tanwīn} symbols when combined with \textit{alif} were standardised to the modern style, so the diacritic symbol appears on top of the \textit{alif} and not on the letter preceding it.

\subsubsection{\textit{Beliefs and Opinions}}  
{We identified} an additional parallel text, namely  Sa'adiah Gaon's  \textit{The Book of Beliefs and Opinions} (\textit{Kitāb al-Amānāt wa l-I`tiqādāt}). 
It contains roughly the same quantity of data and was also transliterated by Bashir. 
We did not use it as training data; 
rather, we extracted 20\% of it as additional test data, for evaluating how well the model performs on unseen text.

\subsubsection{Synthetic data}\label{data_synt}
Although we have a considerable amount of parallel data, since we use only one text source for training, we  faced the problem of the model fitting to the specific writing style of the training data, but generalizing {poorly} to other texts.
To improve the model and make it more robust, we generated synthetic data, using a simple technique, again leveraging the fact that the opposite direction of transliteration, that is, from Arabic to Hebrew, has almost no ambiguities (at least in the sources that we worked with in this study).

We found some additional texts online that correspond to the same era in which our texts were written. 
\hl{We used a simple mapping to produce a pseudo-transliteration from Arabic to Judeo-Arabic for them (Table {\ref{tab:simp_map}} in Appendix).}
For instance, for the first two words in the text of \textit{Ilāhiyyāt},
    \includegraphics[vsmash=true,width=20mm,height=14pt,vshift=-1mm]{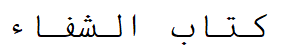}, 
a pseudo-Judeo-Arabic transliteration is generated:
\includegraphics[vsmash=true,width=20mm,height=14pt,vshift=-4pt]{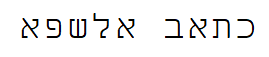}\!.
The generated Judeo-Arabic text along with its Arabic counterpart are added to the parallel data for training the model.

This gave us parallel data that are genuine on the Arabic side but fictitious on the Judeo-Arabic side. 
Thus, it might include words written in a different way than an original Judeo-Arabic author might have written them. 
The  use of such training data is justified partly by the fact that we are likewise only interested in the accuracy of our model’s predictions on the target (Arabic) side. 
Therefore, we are less concerned about providing the model with noisy examples. 
This synthetic data significantly enlarged the quantity available for training.

\section{Baseline Algorithm}

The evaluation metric that we use is simply the average edit distance over all examples in the test dataset. 
The edit distance (ED) that we use is the Levenshtein distance, which is calculated between the predicted characters and the ground truth. It is then normalized by the length of the ground truth.
{The formula for} \textit{label error rate (LER)} is
$\frac{1}{|S|}\sum_{(x,z)\in S}{\textit{ED}(h(x),z)}/{|z|} $
for model $h$ on test data $S\subseteq X\times Z$, where $X$ are the inputs, $z$ is  ground truth and $|z|$ is the length of $z$.
This is a natural measure for a model where the aim is to produce a correct label sequence \cite{graves2006connectionist}.

To evaluate  results, we start by creating a baseline transliteration on the test data that translate each Hebrew letter to the most common letter according to the predefined mapping mentioned earlier (Table \ref{tab:simp_map} in Appendix).
We produced the list manually according to a modern convention for transliterating Arabic into Hebrew.
This simple mapping achieves a relatively high accuracy (LER 9.51\%)  and demonstrates the nature of this Judeo-Arabic transliteration problem. 
Though it is easy to achieve high accuracy, to produce readable text and to be able to confront ambiguities in the text, some language ability is {desirable}. 
The baseline results still do not guarantee fluent reading of the generated target text.

\subsubsection{Common baseline mistakes}
The baseline errors arise mainly from ambiguous letters that have more than a single mapping. 
In what follows, we  enumerate the most prominent ambiguities.

\myparagraph{Transliteration of  Hebrew  \textit{alef}.}
The letter \textit{alef} (\cjRL{'}) most commonly should be transliterated as the Arabic letter \textit{alif} (\RL{A}). This Arabic grapheme usually indicates {an elongated /a/} vowel attached to the preceding consonant. However, it can also sometimes indicate a glottal stop, that is an \textit{alif with hamza on top}  \smash{(\RL{'})}.  As \cite{habash2010Arabic} mentions, Arabic writers often ignore writing the \textit{hamza} (especially with stem-initial \textit{alifs}) and it  is ``de-facto optional".
This will also lead to false negatives for the test data, deciding that the transliteration is an error while in fact it would be accepted by a human reader, unjustifiably increasing LER; see \cite[Section 2.3]{DBLP:journals/corr/RoscaB16}.
A more complex model could hopefully predict the places were \textit{hamza} is required for disambiguation (for instance words with stem-\textit{non}-initial \textit{alifs}). Alternatively, {such} a model would hopefully have a rich enough memory of the Arabic words it has seen, attaching the \textit{hamza} sign for words it has seen in the training data. If indeed there are two legitimate forms, {this} model will also know to disambiguate according to the context, as we train on sequences, that is, words in context.

{Rarer} cases for transliteration of \cjRL{'} are as follows: \textit{hamza} on the line  (\smash{\RL{"'})} even though it is usually not transcribed in Hebrew. 
Also it can mean an \textit{alif maqsura} (\smash{\RL{Y}}) at the end of a word, but \textit{alif maqsura} is usually mapped to the letter \textit{yod}  (\cjRL{y}).
\novocalize
For instance, the 3-letter word \cjRL{wg'}  is transliterated as the 4-letter \smash{\RL{wjA'}}. 
Here it is not clear whether the letter \cjRL{'} corresponds to the \textit{hamza} or to the long vowel \textit{alif} that precedes it, in which case the \textit{hamza} is missing from the transliteration.
In the baseline algorithm, we map \cjRL{'} to the most common transliteration, that is, a non-\textit{hamza alif}. 
Thus we  miss  all the other variations.

\begin{table}[tp!]
  \centering
  \includegraphics[width=0.3\linewidth]{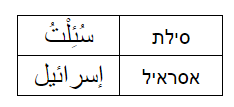}
  \caption[Different transliterations of \RL{Bi'}]{Different transliterations of \RL{Bi'}. In the first row it matches Hebrew \cjRL{y}, while in the second row it matches \cjRL{'}. The unusual transliteration might stem from the spelling of the translation.}
  \label{tab:yod_dif}
\end{table}

\myparagraph{Transliteration of  Hebrew  \textit{yod}.}
The two most common uses of the \textit{yod} (\cjRL{y})  are for the Arabic letter \textit{ya} (\smash{\RL{y}}) and for \textit{alif maqsura} (\smash{\RL{Y}}),  a dotless \textit{ya}  appearing always at the end of a word.
\novocalize
Less frequently it can also be a transliteration of \textit{ya hamza} (\smash{\RL{Bi'}}). 
But there are variations. 
For instance, in the first of the examples in Table \ref{tab:yod_dif}, the \textit{ya hamza} is transliterated as \textit{yod} (\cjRL{y}), while in the second when followed by a regular \textit{ya}, it can be seen as either transliterated to a Hebrew \textit{alef} (\cjRL{'}) or dropped. 
In this example, the variation can be due to the spelling in Hebrew of the translation that uses  \cjRL{'}: “\cjRL{y/sr'l}” (Israel in Hebrew).

In Egypt, but not necessarily in other Arabic countries, a final \textit{ya} is often written dotless, that is, as an \textit{alif maqsura}  \cite{habash2010Arabic}.
This seems to be the case in the transliteration of {Maimonides'} book, \textit{The Guide for the Perplexed} (\textit{Dal\=alat al-\d{h}\=a’ir\={\i}n}; {\small\url{http://sepehr.mohamadi.name/download/DelalatolHaerin.pdf}}), as transliterated by Hussein Attai (e.g.\@ p.\@ 45).
Unfortunately, the book is not available as a digital text. 
(A page of the \textit{Guide}, with human and machine transliterations, is shown in Figure \ref{fig:guide1} in Appendix.)
In the baseline algorithm, we map \textit{yod} invariably to a regular \textit{ya}.

\myparagraph{Arabic \textit{shadda} (gemination).}
Another critical issue is the \textit{shadda} diacritic (e.g.\@ \RL{bb}). 
The shadda is not present in the Judeo-Arabic text, or was omitted in the digitized version of the text as described above. Unfortunately, we could not figure out a simple rule of thumb to handle the presence or absence of shadda that could be adopted for the baseline algorithm.

\begin{figure}[tp!]
  \centering
  \includegraphics[width=0.95\linewidth]{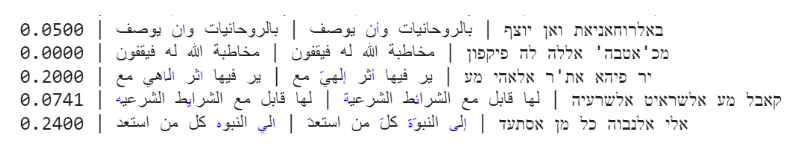}
  \caption[Baseline results]{Baseline results. Columns, right to left: Judeo-Arabic text, ground truth, baseline prediction, error rate.   
 Errors are marked in blue. Observe, for instance, the 4th word in row 3, missing a \textit{shadda} in the prediction and with an extra \cjRL{'} in the source sentence.
}
  \label{fig:base_res}
\end{figure}

\subsubsection{Baseline results}
As mentioned above, the baseline algorithm fully disregards \textit{shadda}, \textit{hamza}, \textit{alif maqsura}, \textit{tanwin} and a few other marks. 
Nonetheless the error on the test data does not {increase} (LER 9.5\%). 
Some results of the baseline algorithm are shown in Figure \ref{fig:base_res}.

\section{Method and Results}

\subsection{Training using CTC Loss with Letter Doubling}\label{CTC}
The connectionist temporal classification (CTC) loss is a method introduced in \cite{graves2006connectionist}  that enables a tagging recurrent neural network (RNN) to learn to predict discrete labels from a continuous signal without requiring the training data to be aligned input to output. 
Instead, the model produces a distribution over all alignments of all possible labels while facilitating an extra character (the \textit{blank} symbol) added to its softmax layer in order to produce the alignment. 
Thus, the probability of any label conditioned on the input can be calculated as the sum over all possible alignments of the given label. 
CTC is {appropriate} for our mission because the Judeo-Arabic inputs and Arabic outputs are not always of equal length. Also, there is no available alignment of the two texts at the character level.

Applying  CTC loss is a convenient solution for handling this problem. 
However, the CTC loss is only defined when the input sequence is longer than the output sequence, which is not always the case for us. 
Actually, since there are diacritic signs in the Arabic transliteration included in the character count that do not appear in the Judeo-Arabic source---this is a prevalent situation. 
This will require an adaptation of the dataset below. 

It should be noted that, as \cite{DBLP:journals/corr/ChanJLV15} describes, the CTC mechanism implicitly assumes independence of characters over time. 
By using the multiplicative rule between probabilities over time, it disregards the dependencies between timestamps. 
In spite of this strong assumption, CTC  achieves a substantial improvement in performance for various sequential tasks, such as speech recognition \cite{graves2006connectionist}, OCR \cite{DBLP:journals/corr/ShiBY15} and handwriting recognition \cite{inproceedings1}.

\subsubsection{Doubling}
This technique of dealing with shorter input than output sequences that we use  is similar to that used by \cite{DBLP:journals/corr/RoscaB16}. 
But instead of using a special character (epsilon) inserted between each timestamps a constant number of times, known as \textit{epsilon insertions}, we filled the extra spaces by repetition of the previous timestamp label. 
For instance, the sequence
\includegraphics[vsmash=true,width=25mm,height=12pt,vshift=-1mm]{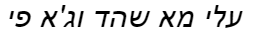}
would become
\includegraphics[vsmash=true,width=50mm,height=14pt,vshift=-1mm]{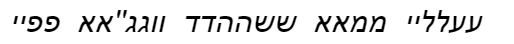}\hspace{-8pt}.
Note the doubling of the {apostrophe} sign,
which is performed separately from the letter \cjRL{g} that it decorates.

Aesthetically speaking, this brings the usage of CTC closer to its original usage of transcribing a continuous signal to discrete labeling (as in speech recognition) in the sense that the input text obtains the appearance of a continuous signal, whereas epsilon insertions  break the succession. 
Further work is required to examine the impact  on performance of these two alternatives.

\subsection{Training the RNN}

We use for our model GRU \cite{cho-etal-2014-learning} cells, stacked in four layers, followed by a linear layer activated by softmax for the CTC loss. 
Each layer is bidirectional (meaning the cells observed the input both backwards and forwards), and contains 1,024 units. 
We use letter embedding for the input of dimension 8. 
The model is implemented with TensorFlow, and optimized with RMSprop. The text is divided into short 20 characters sequences (according to the lengths in the input side). 
The sequences contain complete words. 
If the 20th character happens to be midword, the rest of the word was included in the sequence. The batch size is 128.

\subsection{Pretraining on Single Letters}
A method that was beneficial for speeding up convergence of the network was to pretrain the network with single letters according to the simple letter mapping between the Hebrew and Arabic alphabet. This is as if ``to set the model in the right direction”. This is intuitively reasonable if you think of the way children learn how to read. First they learn to identify the individual letters and then to assemble them into words. We use for this training step the \textit{sparse cross entropy loss} trained on generated random parallel character sequences of length 10. 
{As we  show}, this makes it feasible to train deeper networks. 
It might also be helpful, for instance in the task of speech recognition, to pretrain the network first on single time samples from a certain phoneme to teach the network to map to the correct grapheme, before training on continuous speech with CTC loss, which is more complex, and with which it is less obvious for the network how to start to optimize than with the simpler cross entropy loss.

\begin{figure}[t!]
  \centering
    \subfloat[Training loss: without pretrain] {\includegraphics[width=0.32\textwidth,height=0.2\textwidth]{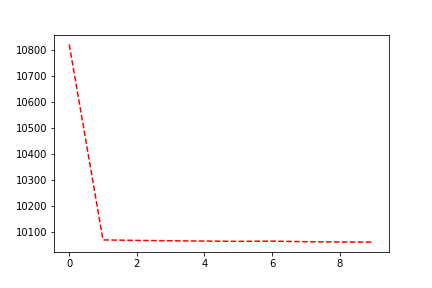}}
  \subfloat[Test loss: without pretrain] {\includegraphics[width=0.32\textwidth,height=0.2\textwidth]{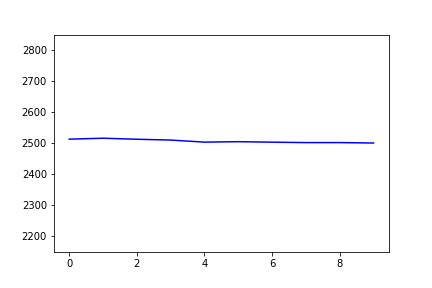}}
  \subfloat[Test accuracy: without pretrain] {\includegraphics[width=0.32\textwidth,height=0.2\textwidth]{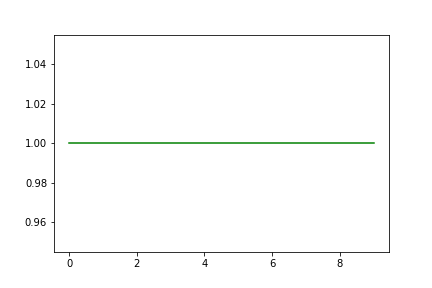}}

    \subfloat[Training loss: with pretrain] {\includegraphics[width=0.32\textwidth,height=0.2\textwidth]{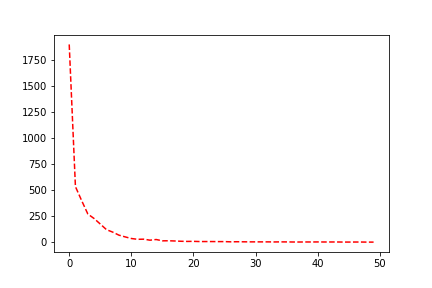}}
  \subfloat[Test loss: with pretrain] {\includegraphics[width=0.32\textwidth,height=0.2\textwidth]{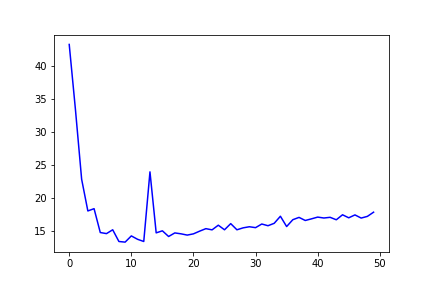}}
  \subfloat[Test accuracy: with pretrain] {\includegraphics[width=0.32\textwidth,height=0.2\textwidth]{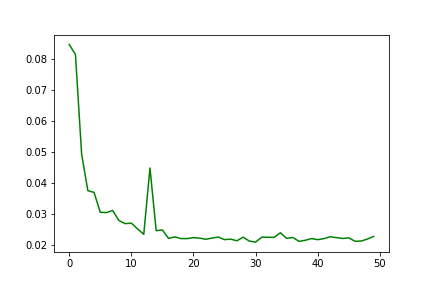}}

  \caption[Comparing results with pretraining on single graphemes 
  and without]{Comparing results with pretraining with single graphemes with CE loss (bottom) and without (top). Beware of the differences in scales.}
  \label{fig:emp_res1}
\end{figure}

\subsubsection{Intermediate results}
With pretraining with the cross entropy loss, the network converges quite fast to reach error rate of 2\% on the test data. 
On the other hand, without pretraining, it gets stuck at a local minimum, transliterating each character in the input to the most prevalent character in the target data, which is a space character, producing as output strings of repeating spaces. Figure \ref{fig:emp_res1} shows the losses and accuracy measures. 
In the remainder of the experiments, we include this pretrain stage. 

Running this model against the additional text of \textit{Emunoth ve-Deoth} yields a higher error rate of 3.24\%. 
In other words, we see a decrease in performance for unseen texts. 
We continue with an exploration of ways for improving prediction for unseen texts.

\subsection{Training with Synthetic Data}
As described in Section \ref{data_synt}, we propose a technique to augment the training data by generating pseudo-parallel texts using Arabic writings of roughly the same era in which our Judeo-Arabic texts were written. 

The texts we used for this purpose in the current experiments are: (1) {Avicenna}'s  \textit{Ilāhiyyāt}, (2) {Al-Farabi}'s \textit{Kitab Rilasa al-Huruf},  (3) {Al-Farabi}'s \textit{Kitab Tahsil al-Saida} and  (4) {Averroes}'s \textit{Al-Darurī fī Isul al-Fiqh}. To all that we add the original ``real'' dataset of the \textit{Kuzari}. By doing this, we significantly enlarge the amount of data that we have for training. 

\subsubsection{Results}
\begin{figure}[tp!]
  \centering
    \subfloat[Training loss] {\includegraphics[width=0.32\textwidth]{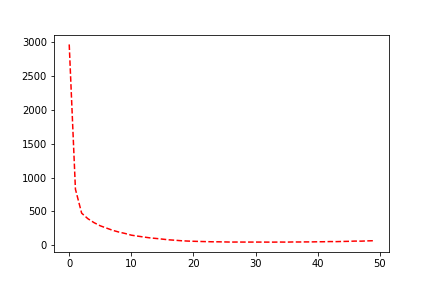}}    
  \subfloat[Test loss] {\includegraphics[width=0.32\textwidth]{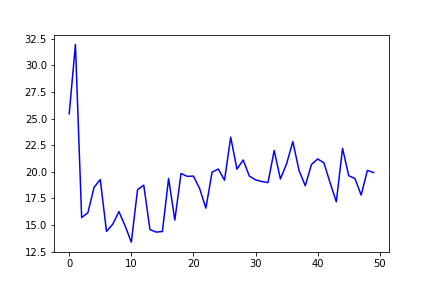}}  
  \subfloat[Test accuracy] {\includegraphics[width=0.32\textwidth]{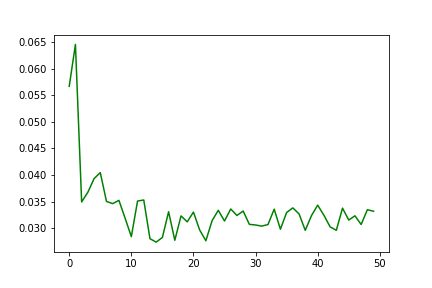}}
  \caption{Results of training with added synthetic data.}
  \label{fig:emp_res2}
\end{figure}

With the synthetic data, the accuracy on the original text data \textit{decreased}, as expected, to 2.48\%. On the additional data the accuracy also \textit{decreased} but to a lesser extent, to 3.37\%. This drop in performance might indicate that the Arabic texts that we chose for generating the synthetic data are not perfectly suitable for Judeo-Arabic.
Perhaps there are other sources to consider that are more suitable.
Note that the degree of fluctuation in the learning curve on the test is greater. This might be because the synthetic data is shuffled between epochs. 
The results are presented in Figure \ref{fig:emp_res2}.

\subsection{Dropout}
We are mainly interested in making the network to remember what would make sensible Arabic outputs, relying on the letter mapping, but allowing some flexibility. 
It should be able to output Arabic words that were seen during  training  even though the input sequence is not the exact one seen during  training. {Experimenting} with a small model showed the plausibility of this direction of thinking.
An immediate implication was to use \textit{dropout} on the input sequence to make the network more robust to noise in the Judeo-Arabic input. We achieved this by randomly setting a percentile of the nonspace symbols in the input sequence (before doubling the letters) to the \textit{blank} symbol.

\subsubsection{Results}
With dropout rate of 15\% of the nonspace symbols that was performed on the synthetic data, the test results for the original test of the \textit{Kuzari} is set at 2.26\%, which is worse than the accuracy without synthetic data and without dropout but better than the results when trained on the synthetic data without the dropout.
On the other hand, for the additional text of \textit{Emunot}, a {marked improvement is achieved, yielding} an error rate of 3.14\% (a 0.1 percent improvement).

\section{Discussion}

\begin{figure}[t!]
\centering
\begin{description}
\item[(a)] {\includegraphics[width=0.95\textwidth,vshift=-2pt,height=10pt]{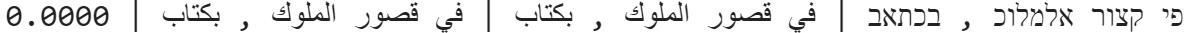}}
       \item[(b)] {\includegraphics[width=0.95\textwidth,vshift=-3pt,height=12pt]{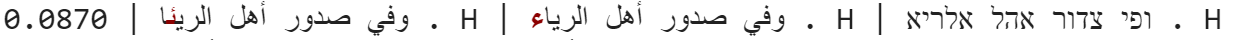}}
       \item[(c)] {\includegraphics[width=0.95\textwidth,vshift=-4pt,height=12pt]{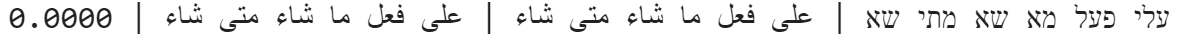}}
       \item[(d)] {\includegraphics[width=1.0\textwidth,vshift=-2pt,height=10pt]{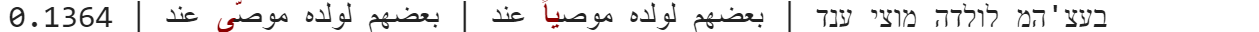}}
       \item[(e)] {\includegraphics[width=0.95\textwidth,vshift=-3pt,height=12pt]{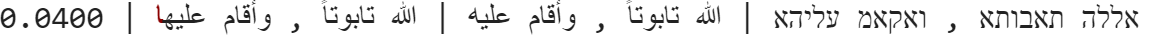}}
   \item[(f)] {\includegraphics[width=0.95\textwidth,vshift=-3pt]{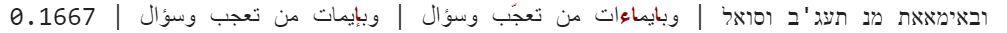}}
   \item[(g)] {\includegraphics[width=0.85\textwidth,vshift=-2pt,height=12pt]{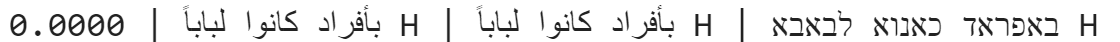}}
   \item[(h)] {\includegraphics[width=0.95\textwidth,vshift=-4pt,height=14pt]{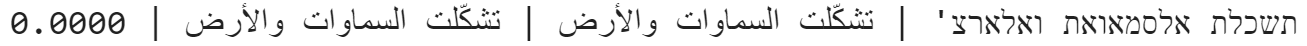}}
   \item[(i)] {\includegraphics[width=0.95\textwidth,vshift=-2pt]{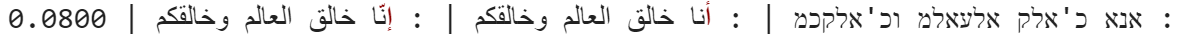}}
   \item[(j)] {\includegraphics[width=0.95\textwidth,vshift=-4pt]{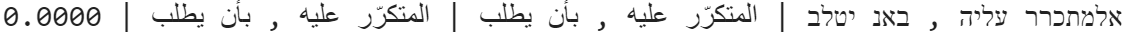}}
    \end{description}
    \caption[Examples illustrating transliteration results for proposed model]{Example transliteration results for proposed model. Right to left: Judeo-Arabic, ground truth, baseline prediction, error rate.
    In {(b)}, for the last word the network misses the ground truth. 
    Notice the correct positioning of the \textit{Alif Hamza}.
In {(d)} the mistake is partly due to noise in the data since the ground truth form \newtanwin
\smash{\RL{m_U.sy"aN}} is a conjugation of the noun \smash{\RL{m_U.sy}} (meaning ``recommender''), which the Judeo-Arabic \smash{\cjRL{mw.sy}} form  resembles more. 
The form 
that the network finally produced also exists.
Example {(i)} shows a mistake in distinguishing between word senses since both diacritizations of \smash{\RL{anA}} produce legitimate words.
    }
    \label{fig:results_disc}
\end{figure}

We {designed} a model for automatic transliteration of Judeo-Arabic texts into Arabic scripts. Endeavoring to overcome the problem of ambiguous mappings in the transliteration, we trained an RNN model using the CTC loss that has enabled us to cope with unequal length input/output sequences due to the addition of diacritics in the target (Arabic) side of the data. As mentioned, we wanted to create a network that will have memory, along with some language capability in the output side that will enable it to distinguish between different word senses and overcome small variations in the transliteration. 

The results demonstrate a substantial decrease in error compared to the baseline, from 9.5\% to 2\% LER, showing that the network is capable of attaching correct diacritics to the Arabic.
On unseen text, the network {incurs} a 3.24\% error rate. 
This is important since Judeo-Arabic texts vary according to the writer,  time period, and  region.
Examples {and some remaining problems} are included in Figure \ref{fig:results_disc}.

We experimented with several methods to enhance  training. 
First, we exercised pre-training of the network with cross-entropy 
 loss to teach the network the simple mapping used in the baseline transliteration. This can enlarge and deepen the network and still guarantee convergence. 
 Without this pretraining stage, the network failed to discover this mapping by itself.
Since we only train on parallel text from a single source, and we are interested in making the model generalize better to unseen texts, we augment the training dataset by generating parallel texts out of available medieval Arabic texts.
This had a slight adverse affect on accuracy and might depend on the choice of Arab texts that were used for the synthetic data.
Dropout was considered and implemented on the synthetic data with the desired affect. 
While accuracy on the first test data, taken from the same source as the original training data decreased as expected,  accuracy for the additional test data improved. 
We suspect that the network assimilates more language skills due to the dropout on the training data.


Examining the top 5 results of the CTC-decode beam search, instead of only the top one, reveals that sometimes the correct transliteration, at least the one that was chosen by the human transliterator, is among those five. 
Therefore, running a pure Arabic language model should enhance model accuracy.

One question that arises is how much of the language the network catches. 
To examine this, we perform a forward pass on the shuffled test data. 
Shuffling is done at word level. 
This generates a parallel dataset of words that lack context, {sharing} the same word distribution as  the real test data. Indeed as expected, on the shuffled test data the error increases by $\sim$0.5\% on average.

\begin{ignore}
Since we transliterate sequences of several words, we utilize the context of each word in predicting its correct transliteration. This is available only for words that are not the first and last in the sequence. We propose to change the setting of the testing to regard only words inside  a fixed context window, and to calculate the loss and accuracy only on words that have context. We suspect this will reveal that errors occur more frequently at the start or end of a sequence.
Another plan will be to test only on a single middle word inside a constant size context window. 
It will be interesting to check how results change with different window sizes, including a window size of \textit{zero}, which will be another way to check how much of the language skills the model learns, a model that is not using explicitly a language model.
\end{ignore}

\newpage
\bibliographystyle{coling}
\bibliography{references}

\newpage
\begin{appendix}
\section*{Appendix: Figures and Tables}


\begin{figure}[htp!]
    \centering
    \subfloat[Original Judeo-Arabic orthography.]{
    \includegraphics[width=0.48\textwidth]{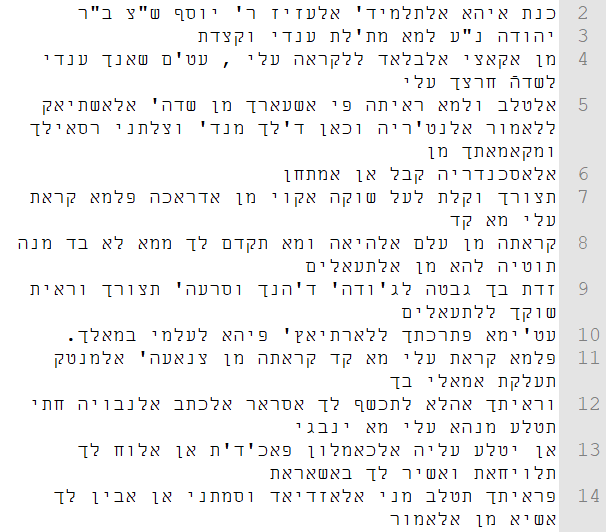}}
    \subfloat[Transliteration by {Hussein Attai}.]
    {\includegraphics[width=0.55\textwidth]{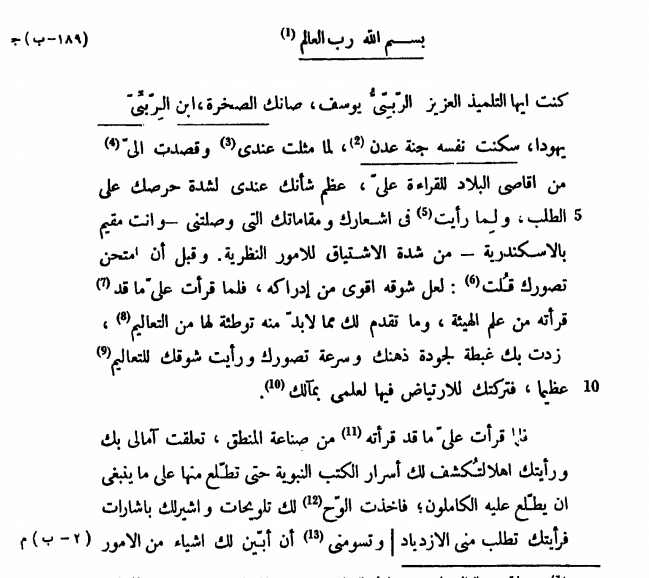}}

    \subfloat[Our model transliteration.]
    {\includegraphics[width=0.8\textwidth]{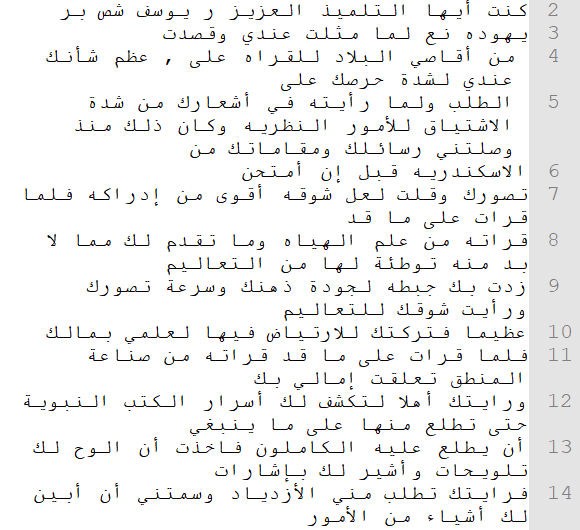}}
        \caption{First page of {Maimonides}' \textit{The Guide for the Perplexed}.}
      \label{fig:guide0}\label{fig:guide1}
    \label{fig:guide}
\end{figure}

\begin{table}[htp!]
  \caption{Simple mapping rules for baseline transliteration.}
  \label{tab:simp_map}
  \centering
  \includegraphics[width=0.5\linewidth]{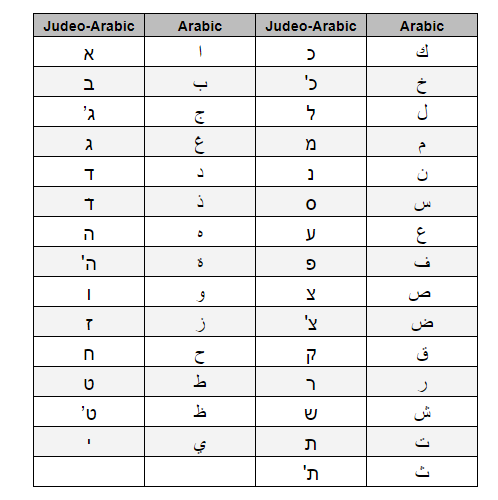}
\end{table}

\end{appendix}
\end{document}